\documentclass[a4paper, 10pt, conference]{ieeeconf}

\newif\ifarwfinalcopy

\arwfinalcopytrue  % *** Uncomment this line for the final submission
\IEEEoverridecommandlockouts                       

\usepackage{graphics} % for pdf, bitmapped graphics files
\usepackage{epsfig} % for postscript graphics files
\usepackage{mathptmx} % assumes new font selection scheme installed
\usepackage{times} % assumes new font selection scheme installed
\usepackage{amsmath} % assumes amsmath package installed
\usepackage{amssymb}  % assumes amsmath package installed
\usepackage{booktabs}
\usepackage{multirow}

\newcommand{\cmark}{\textcolor{green}{\checkmark}} % Define green checkmark
\newcommand{\xmark}{\textcolor{red}{\texttimes}} % Define red cross

\usepackage{algorithmic}
\usepackage{graphicx}
\usepackage{textcomp}
\usepackage{xcolor}

\usepackage{lineno}
\usepackage{tikzpagenodes}
\usepackage{background}

\ifarwfinalcopy
\backgroundsetup{color=white}
\else

\linenumbers

\newcommand{\MyARWConfidentialLogo}{% For a logo drawn with TikZ
\begin{tikzpicture}[remember picture,overlay]
\node[align=center,text=blue] at ([yshift=1em]current page text area.north) {\Large \#\#\# ARW 2025 SUBMISSION: CONFIDENTIAL REVIEW COPY \#\#\#};
\end{tikzpicture}%
}

\SetBgContents{\MyARWConfidentialLogo}% Set tikz picture
\SetBgPosition{current page.north west}% Select location
\SetBgOpacity{0.5}% Select opacity
\SetBgAngle{0.0}% Select rotation of logo
\SetBgScale{1.0}% Select scale factor of logo

\fi

\title{\LARGE \bf
Category-Level and Open-Set Object Pose Estimation for Robotics}

\author{Peter Hönig, Matthias Hirschmanner, and Markus Vincze% (do not remove)
\thanks{All authors are with the Automation and Control Institute, Faculty of Electrical Engineering, TU Wien, 1040 Vienna, Austria {\tt\small \{hoenig, hirschmanner, vincze\}@acin.ac.tuwien.at}}%
}

\begin{document}

\maketitle

%%%%%%%%%%%%%%%%%%%%%%%%%%%%%%%%%%%%%%%%%%%%%%%%%%%%%%%%%%%%%%%%%%%%%%%%%%%%%%%%
\begin{abstract}
Object pose estimation enables a variety of tasks in computer vision and robotics, including scene understanding and robotic grasping.
The complexity of a pose estimation task depends on the unknown variables related to the target object.
While instance-level methods already excel for opaque and Lambertian objects, category-level and open-set methods, where texture, shape, and size are partially or entirely unknown, still struggle with these basic material properties.
Since texture is unknown in these scenarios, it cannot be used for disambiguating object symmetries, another core challenge of 6D object pose estimation.
The complexity of estimating 6D poses with such a manifold of unknowns led to various datasets, accuracy metrics, and algorithmic solutions.
This paper compares datasets, accuracy metrics, and algorithms for solving 6D pose estimation on the category-level.
Based on this comparison, we analyze how to bridge category-level and open-set object pose estimation to reach generalization and provide actionable recommendations.
\end{abstract}


%%%%%%%%%%%%%%%%%%%%%%%%%%%%%%%%%%%%%%%%%%%%%%%%%%%%%%%%%%%%%%%%%%%%%%%%%%%%%%%%
\section{INTRODUCTION}
\begin{figure}[t!]
   \centering
    \includegraphics[width=1.0\columnwidth]{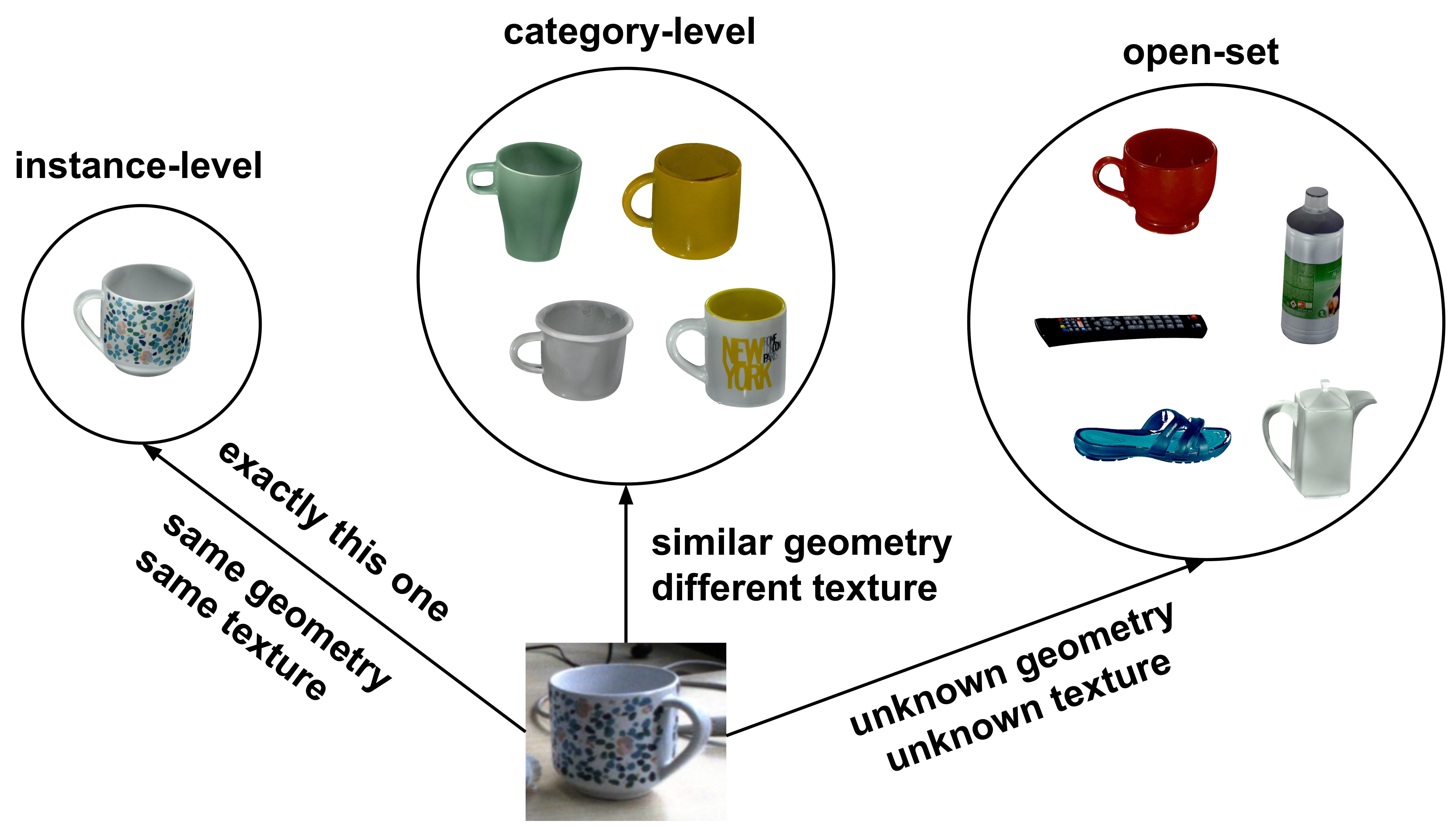}
   \caption{\textbf{Comparison of instance-level, category-level, and open-set object pose estimation.} The complexity of the pose estimation task is increasing from left (instance-level) to right (open-set) due to the number of unknowns.}
   \label{fig:comparison}
\end{figure}

Object pose estimation is necessary in robotics for tasks such as robotic grasping~\cite{thalhammer2024challenges}.
If the geometry of a target object is known, its 6D pose, defined by the rotation $\mathbf{R}$ and translation $\mathbf{t}$, is sufficient to locate it in a $SE$(3) space.
This definition is insufficient as soon as the target object geometry is only partially known, as in the case of category-level object pose estimation.
Open-set object pose estimation is even more complex than category-level object pose estimation since both geometry and texture are entirely unknown.
In category-level and open-set object pose estimation, a sole 6D pose leaves unknowns to describe the object for tasks such as grasping.
In these cases, additional information is necessary, such as a 7D pose ($\mathbf{R}$, $\mathbf{t}$, and $s$ for scale), a 9D pose ($\mathbf{R}$, $\mathbf{t}$, and $\mathbf{s}$, a 3D vector with $x,y,z$ dimensions of the aligned bounding box), or a 6D pose in combination with a shape reconstruction.
The differences between instance-level, category-level, and open-set object pose estimation are illustrated in Fig. ~\ref{fig:comparison}.
The three circles in Fig. ~\ref{fig:comparison} represent the prior knowledge available to a pose estimation algorithm during a training or onboarding stage.
During the inference stage existing category-level algorithms~\cite{wang2019normalized, ikeda2024diffusionnocs, wan2023socs} do not require a 3D object model.
However, in open-set object pose estimation, recent methods do require an object model during inference~\cite{ausserlechner2024zs6d}, or they reconstruct an object mesh from multiview RGB images during an onbording stage~\cite{ornek2024foundpose}.
These reconstructions however are prone to reconstruction noise and the full object surface needs to be visible in order for a full reconstruction.
This does not allow one-shot pose estimation, where only a single frame is available.
% While object detection (localization of an object in 2D) works well on the category-level, pose estimation struggles with the variety in geometry and texture.

% Recently published algorithms \cite{wang2021gdrnet, su2022zebrapose} designed to solve 6D pose estimation on the instance level have led to saturated accuracy metrics on benchmarking datasets~\cite{sundermeyer2023bop}.
% Therefore, the focus in the instance-level domain has shifted toward improving 6D pose estimation of textureless, metallic, and transparent objects~\cite{thalhammer2024challenges, hoenig2024star}.
% Beyond object instances, estimating the 6D pose for whole object categories, with variation in texture and geometry, remains challenging.
In order to make 3D object models during inference obsolete, knowledge has to be induced during a training stage.
Consequently, canonicalization is required.
Canonicalization describes the centering and alignment of objects canonically in the $SE$(3) space.
To address the issues of texture and geometry variation, a color encoding of object geometry is proposed in~\cite{wang2019normalized}, namely the Normalized Object Coordinate Space (NOCS).
NOCS describes a dimensionless $1\times1\times1$ cube, where the $x,y,$ and $z$ coordinates of canonically oriented objects are mapped to RGB values.
This geometrical color encoding is used in other category-level and open-set object pose estimation solutions~\cite{tian2020shape, lee2021noce, wang2021recurrent, zhang2022ssp, zou2023gpt, ikeda2024diffusionnocs, krishnan2024omninocs}, becoming the de-facto standard intermediate representation in the field.

\begin{figure*}[t!]
   \centering
    \includegraphics[width=2.0\columnwidth]{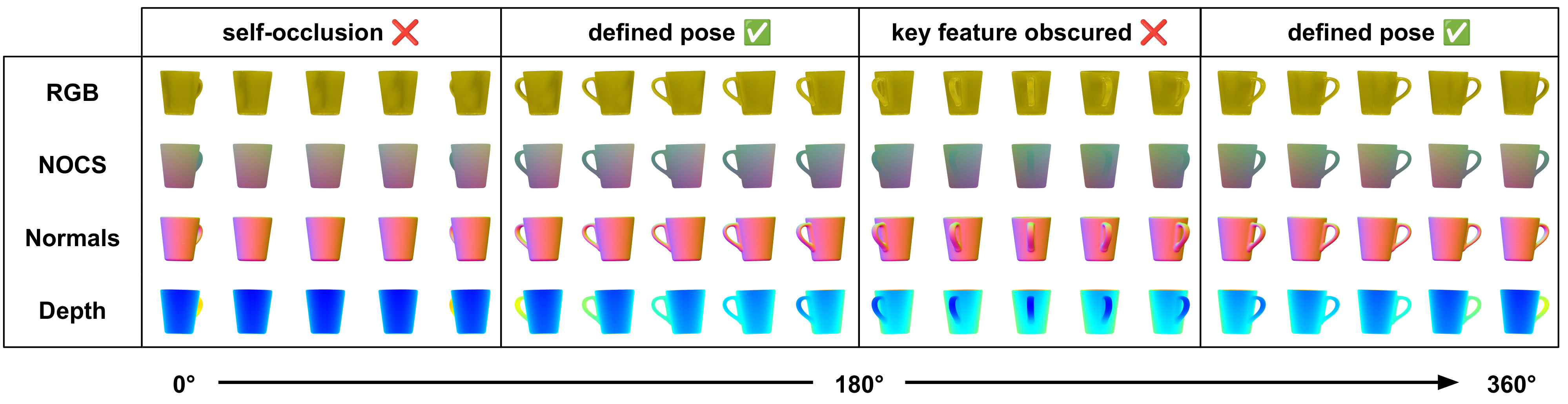}
   \caption{\textbf{Comparing Input Modalities.} A mug is rotated 360$^{\circ}$ around the y-axis to showcase how a key feature (the handle) is self-occluded in a 90$^{\circ}$ range, clearly exposed in a 180$^{\circ}$ range, and obscured due to the uniform texture in a 90$^{\circ}$ range. The various input modalities are shown to highlight the mug handle stronger (depth, normals) or weaker (RGB, NOCS).}
   \label{fig:modalities}
\end{figure*}

Besides dealing with texture and shape variations, properly disambiguating object symmetries remains challenging in category-level object pose estimation.
Not accounting for object symmetries may prohibit optimization algorithms from converging correctly.
While NOCS does disambiguate geometrically symmetric objects with distinct textures, it does not account for textureless objects and pointcloud-only input data.
While instance-level object pose estimation uses distinct textures to disambiguate geometric symmetries, category-level object pose estimation deals with variation in object texture.
This texture variety is handled differently during training of neural networks for category-level object pose estimation~\cite{zou2023gpt, ikeda2024diffusionnocs}.

This comparative study analyzes category-level object pose estimation papers~\cite{wang2019normalized, tian2020shape, lee2021noce, chen2021fs, wang2021recurrent, zhang2022ssp, wan2023socs, wang2023query6dof, zou2023gpt, ikeda2024diffusionnocs, fan2024acrpose, krishnan2024omninocs} focusing on symmetry handling and how to bridge category-level and open-set object pose estimation.
We limit our investigation to algorithms predicting the object pose from a single frame, with no 3D object model available during inference.
We only consider algorithms which are evaluated either on the CAMERA~\cite{wang2019normalized} or REAL275~\cite{wang2019normalized} dataset.
Algorithms that use stereo vision~\cite{chen2023stereopose} or consecutive frames~\cite{zaccaria2023opticalflow} or are solely evaluated on other datasets are not included in this study.
While previous works summarize and review the state of the art in object pose estimation in general~\cite{thalhammer2024challenges, marullo20236d, fan2022deep}, we explicitly focus on category-level object pose estimation and how to bridge this technique to the open-set domain, without requiring 3D object models during the inference step.
Our contributions can be summarized as follows:
\begin{itemize}
    \item A concise review of the state of the art in category-level object pose estimation with an emphasis on input modalities, network architectures, 6D pose solvers, and rotational symmetry handling.
    \item Actionable recommendations for tailoring future category-level object pose estimation methods for generalization beyond known categories to bridge to the open-set domain.
\end{itemize}

The following section starts by discussing types of input modalities.
We will discuss how input modalities influence symmetry disambiguation.
The subsequent section reviews network architectures.
In this section we will discuss how network architectures can handle symmetric objects differently and how network types relate to model performances.
Next, we discuss types symmetry handling, their advantages and limitations.
We continue with a section about 6D pose solvers, discussing deterministic and learned variants and we elaborate on pose estimation performance metrics used for comparison.
We conclude with experiments and results, discussing current performances of category-level object pose estimation algorithms and give actionable recommendations for potential improvements based on the findings during our literature research.
The papers are compared in Table \ref{tab:pose_estimation_comparison}.
%The following section will begin the comparison by discussing different input modalities.

\section{INPUT MODALITIES}
\label{sec:input_modalities}
The choice of input modalities influences how symmetries are resolved~\cite{thalhammer2024challenges}.
For specific object geometries, RGB-only leads to ambiguous views, where key features are not clearly exposed, e.g., when the handle of a textureless mug directly points toward the camera.
This phenomenon is depicted in Fig.~\ref{fig:modalities}.
Depth input provides complementary geometric cues, which resolves such ambiguities.
Furthermore, since texture and shape are varying in category-level object pose estimation, depth or normals input provide actual geometric surface information.
Since category-level object pose estimation algorithms do not access 3D object meshes during the inference stage, algorithms using depth or normals are in favor.
In robotic applications, depth data is abundant~\cite{thalhammer2024challenges, bauer2020verefine} making algorithms that use depth modality the preferred option.

The papers~\cite{wang2019normalized, tian2020shape, lee2021noce, wang2021recurrent, ikeda2024diffusionnocs, fan2024acrpose, krishnan2024omninocs} use RGB input for their algorithms, as listed in Table~\ref{tab:pose_estimation_comparison}.
\cite{wang2019normalized, tian2020shape} use RGB for establishing 2D-3D correspondences and depth for transforming normalized 3D to metric 3D coordinates.
\cite{wang2021recurrent, fan2024acrpose} use RGB and depth for predicting 2D-3D correspondences and depth again for 6D pose solving between normalized and metric 3D.
\cite{chen2021fs, zhang2022ssp, wan2023socs, wang2023query6dof, zou2023gpt} rely on depth only for pose estimation and use RGB for the object detection stage to estimate region proposals.
\cite{lee2021noce, krishnan2024omninocs} use solely RGB data, while \cite{krishnan2024omninocs} extracts DINOv2 features from RGB before processing further.
\cite{ikeda2024diffusionnocs} use RGB, depth, and DINO features.

\begin{table*}[t!]
\centering
\caption{\textbf{Comparison of Category-Level Object Pose Estimation Papers.} Papers presenting algorithms predicting category-level object pose from single frames.}
\label{tab:pose_estimation_comparison}
\begin{tabular}{l|c|cc|c|c|c|c}
\toprule
\multirow{2}{*}{\textbf{Year}} & \multirow{2}{*}{\textbf{Paper}} & \multicolumn{2}{c|}{\textbf{Input}} & \multirow{2}{*}{\textbf{NOCS}} & \multirow{2}{*}{\textbf{Symmetry Handling}} & \multirow{2}{*}{\textbf{Network}} & \multirow{2}{*}{\textbf{6D Pose Solver}} \\
 & & RGB & Depth & & & & \\
\midrule
2019 & Wang et al.~\cite{wang2019normalized} &  \cmark & \cmark & \cmark & sym. transform loss~\cite{pitteri2019object} & Mask R-CNN like & Umeyama~\cite{umeyama1991least}\\

 2020 & Tian et al.~\cite{tian2020shape} &\cmark & \cmark & \cmark & sym. transform loss~\cite{pitteri2019object} & Encoder-decoder CNN & Umeyama~\cite{umeyama1991least}\\

2021 & Lee et al.~\cite{lee2021noce} & \cmark & \xmark & \cmark & none & Encoder-decoder CNN &  Umeyama~\cite{umeyama1991least}\\

2021 & Chen et al.~\cite{chen2021fs} &  \xmark & \cmark & \xmark & sym. transform loss~\cite{pitteri2019object} & Encoder-decoder CNN & Direct regression\\

2021 & Wang et al.~\cite{wang2021recurrent} &  \cmark & \cmark & \cmark & none & Recurrent reconstruction CNN &  Umeyama~\cite{umeyama1991least}\\

2022 & Zhang et al.~\cite{zhang2022ssp} &  \xmark & \cmark & \cmark & sym. transform loss~\cite{pitteri2019object} & 3D GCN & Direct regression\\

2023 & Wan et al.~\cite{wan2023socs} &  \xmark & \cmark & (\cmark) & none & 3D GCN &  Anisotropic scaling\\

2023 & Wang et al.~\cite{wang2023query6dof} &  \xmark & \cmark & \cmark & none & CNNs + MLPs &  Direct regression\\

2023 & Zou et al.~\cite{zou2023gpt} & \xmark & \cmark & \cmark & sym. transform loss~\cite{pitteri2019object} & Transformer & Umeyama~\cite{umeyama1991least}\\

2023 & Remus et al.~\cite{remus2023icnet} & \cmark & \cmark & \xmark & none & Encoder-decoder CNN & Direct regression \\
2024 & Ikeda et al.~\cite{ikeda2024diffusionnocs} &  \cmark & \cmark & \cmark & probabilistic & Diffusion model & TEASER++~\cite{yang2020teaser} \\

2024 & Fan et al.~\cite{fan2024acrpose} &  \cmark & \cmark & \cmark & none & Encoder-decoder CNN & Umeyama~\cite{umeyama1991least}\\

2024 & Krishnan et al.~\cite{krishnan2024omninocs} & \cmark & \xmark & \cmark & none & Transformer & Direct regression \\

\bottomrule
\end{tabular}
\end{table*}

\section{NETWORK ARCHITECTURES}
\label{sec:network_architectures}
Network architectures vary between the selected papers.
As Table~\ref{tab:pose_estimation_comparison} indicates, recent advancements in computer vision research are reflected in the development of novel category-level object pose estimation methods.
While traditional CNN architectures dominated from 2019 to 2021, Graph Convolutional Networks (GCNs)~\cite{kipf2017semi} were adopted in 2022.
~\cite{ikeda2024diffusionnocs} use diffusion in 2024, and ~\cite{zou2023gpt, krishnan2024omninocs} use transformer models in 2023 and 2024.
When comparing pose estimation accuracy in the subsequent chapters, one must consider that GCNs, transformers, and diffusion models need extended training and inference times compared to CNNs due to their different network characteristics.
This is due to transformers processing the input in multiple heads~\cite{dosovitskiy2020image}, diffusion models requiring multiple denoising steps~\cite{ho2020denoising} and GCNs dealing with not only 2D but 3D data.

\section{6D POSE SOLVER}
\label{sec:6dposesolver}
None of the selected papers directly regress the 6D, 7D, or 9D pose from the input without intermediate steps.
Regressing object poses without intermediate representations was shown to be inefficient~\cite{zhou2019continuity}.
All selected papers first predict an intermediate representation and solve the 6D pose subsequently.
The papers~\cite{wang2019normalized, tian2020shape, lee2021noce, wang2021recurrent, zhang2022ssp, zou2023gpt, wang2023query6dof, ikeda2024diffusionnocs, krishnan2024omninocs} use NOCS as intermediate representation while~\cite{wan2023socs} use an adapted version of NOCS, namely the Semantically-aware Object Coordinate Space (SOCS), a representation similar to NOCS with additional parameters to highlight semantically meaningful regions around keypoints.
\cite{chen2021fs} perform pointcloud reconstruction and regress $\mathbf{R}$, $\mathbf{t}$, and~$s$ directly from latent features of the encoder-decoder network.
% \cite{wang2023query6dof} learn sparse shape queries 

For networks that estimate the NOCS (or SOCS) pointcloud $\mathbf{P}_N$ in normalized space $N$, a final step for transforming $\mathbf{P}_N$ to the metric space $M$ is needed to acquire $\mathbf{P}_M$.
This step involves solving the equation: $\mathbf{P}_{M} = s \cdot \mathbf{R} \cdot \mathbf{P}_{N} + \mathbf{t}_{M}$, where $\mathbf{R}$, $\mathbf{t}_{M}$, and $\mathbf{s}$ are unknown.
The authors of \cite{wang2019normalized, tian2020shape, lee2021noce, wang2021recurrent, zou2023gpt, fan2024acrpose} use the Umeyama algorithm to solve for $\mathbf{R}$, $\mathbf{t}_{M}$, and $\mathbf{s}$, and use metric depth data from a sensor to acquire $\mathbf{P}_{M}$.
\cite{lee2021noce} uses a Metric Scale Object Shape (MSOS) branch  to estimate a metric 3D model parallel to predicting $\mathbf{I}_N$.
Therefore, they predict both $\mathbf{I}_N$ and $\mathbf{P}_{M}$ without using depth data from a sensor.
The independence from depth comes with the drawback of increased runtime and limited performance since the MSOS branch can only interpolate between object models encountered during training.
Objects with measurements beyond the ones seen in the training data (e.g., a realistic model car, vastly smaller than an actual car but with the same semantic properties) will lead to wrong results.
\cite{ikeda2024diffusionnocs} use the TEASER++ algorithm to solve for $\mathbf{R}$, $\mathbf{t}_{M}$, and $s$, and use depth data from a sensor to acquire $\mathbf{P}_{M}$.
\cite{krishnan2024omninocs} directly regress $\mathbf{R}$ from $\mathbf{I}_N$, and regress $\mathbf{t}_{M}$ and $\mathbf{s}$ after the Dense Prediction Transformer (DPT) backbone~\cite{ranftl2021vision}.
\cite{krishnan2024omninocs} predict NOCS from a full scene with full semantic context, which helps to learn metric object size without depth data.
Still, object sizes outside of the distribution seen during training will be challenging for such a model, similar to~\cite{lee2021noce}.
In regards to generalization, the deterministic algorithms Umeyama and TEASER++ do have the advantage of object category agnosticity.
While~\cite{krishnan2024omninocs} performs direct regression for 6D pose solving, added depth and a deterministic pose solver would most likely result in improved performance.

\section{ROTATIONAL SYMMETRY HANDLING}
\label{sec:rotational_symmetry_handling}
While the pose of the textured object is distinct in all four views, the pose of the textureless object is not.
The selected papers of this comparative study handle symmetry differently.
Besides no explicit symmetry handling, three major symmetry handling techniques are used.

% \begin{figure}[htbp]
%    \centering
%     \includegraphics[width=1.0\columnwidth]{sym_image.jpg}
%    \caption{\textbf{Illustration of Rotational Symmetries.} A box-shaped object is shown in three configurations. The first row indicates the NOCS formulation of the object, rotated by the three discrete symmetries around x,y, and z. The middle row shows the box-shaped object with a distinct texture, allowing for disambiguation in all symmetrical poses. This e.g., allows for disambiguating between the front and the back of a packaging. The last row shows the box-shaped object in the same symmetry poses but as a textureless object, which does not allow for disambiguation.}
%    \label{fig:sym_image}
% \end{figure}

\subsection{No explicit symmetry handling}
\cite{lee2021noce, wang2021recurrent, wan2023socs, wang2023query6dof, fan2024acrpose, krishnan2024omninocs} do not mention any implicit nor explicit symmetry handling technique.
Pose estimation performance is reported for the whole dataset, not single object categories.
% Therefore, the improvement for non-symmetric objects outweighs the drawback of no symmetry handling.
% The authors of \cite{krishnan2024omninocs} do not use any explicit symmetry handling and state that handling symmetry is an open problem that might be addressed in their future research.

\subsection{Orthogonal vectors}
\cite{chen2021fs} do not predict $\mathbf{R}$ as a 3x3 matrix but instead use two decoders that estimate two perpendicular vectors to describe~$\mathbf{R}$.
For objects with continuous symmetries around one axis the loss weight for one of the two perpendicular vectors is set to 0.
This eliminates the constraint of predicting a fully defined pose, but is only applicable to continuous symmetries.
Since this form of symmetry handling is explicit, symmetry types have to be manually annotated.
For handling discrete symmetries~\cite{chen2021fs} use the symmetry transform loss~\cite{pitteri2019object} described in the following section.

\subsection{Symmetric transform loss}
To handle discrete and continous symmetries~\cite{wang2019normalized, tian2020shape, chen2021fs,zou2023gpt,zhang2022ssp} use the symmetric transform loss described by~\cite{pitteri2019object}.
The symmetric transform loss \( \mathcal{L}_{\text{sym}} \) can be defined as:
\[
\mathcal{L}_{\text{sym}} = \min_{\mathbf{R} \in \mathcal{S}} \mathcal{L}\left(\mathbf{P}_{\text{est}}, \mathbf{R} \cdot \mathbf{P}_{\text{gt}}\right),
\]
where:
\begin{itemize}
    \item \( \mathbf{P}_{\text{est}} \) is the estimated NOCS point cloud.
    \item \( \mathbf{P}_{\text{gt}} \) is the ground truth NOCS point cloud.
    \item \( \mathcal{S} = \{ \mathbf{R}_1, \mathbf{R}_2, \ldots, \mathbf{R}_n \} \) is the set of all rotational symmetry transformations.
    \item \( \mathcal{L}(\mathbf{P}_1, \mathbf{P}_2) \) is the loss function measuring the distance between two point clouds.
\end{itemize}

The symmetry transform loss requires handcrafted annotations of symmetrical poses for all objects in the training set.
Annotation requires the symmetries to be discrete (e.g., a rotational symmetry has to be divided into $n$ discrete symmetries around that axis).
Furthermore, since real-life objects are either reconstructed or modeled with CAD, an arbitrary choice of symmetry is necessary.
Reconstructed meshes are never fully symmetric due to reconstruction noise, and CAD parts may break symmetry only by a minor part of the object in relation to the full size. 
While these alterations technically break the symmetry, defining them as symmetric may still lead to better convergence when predicting NOCS.

\subsection{Probabilistic symmetry handling}
\begin{figure}[t]
   \centering
    \includegraphics[width=0.8\columnwidth]{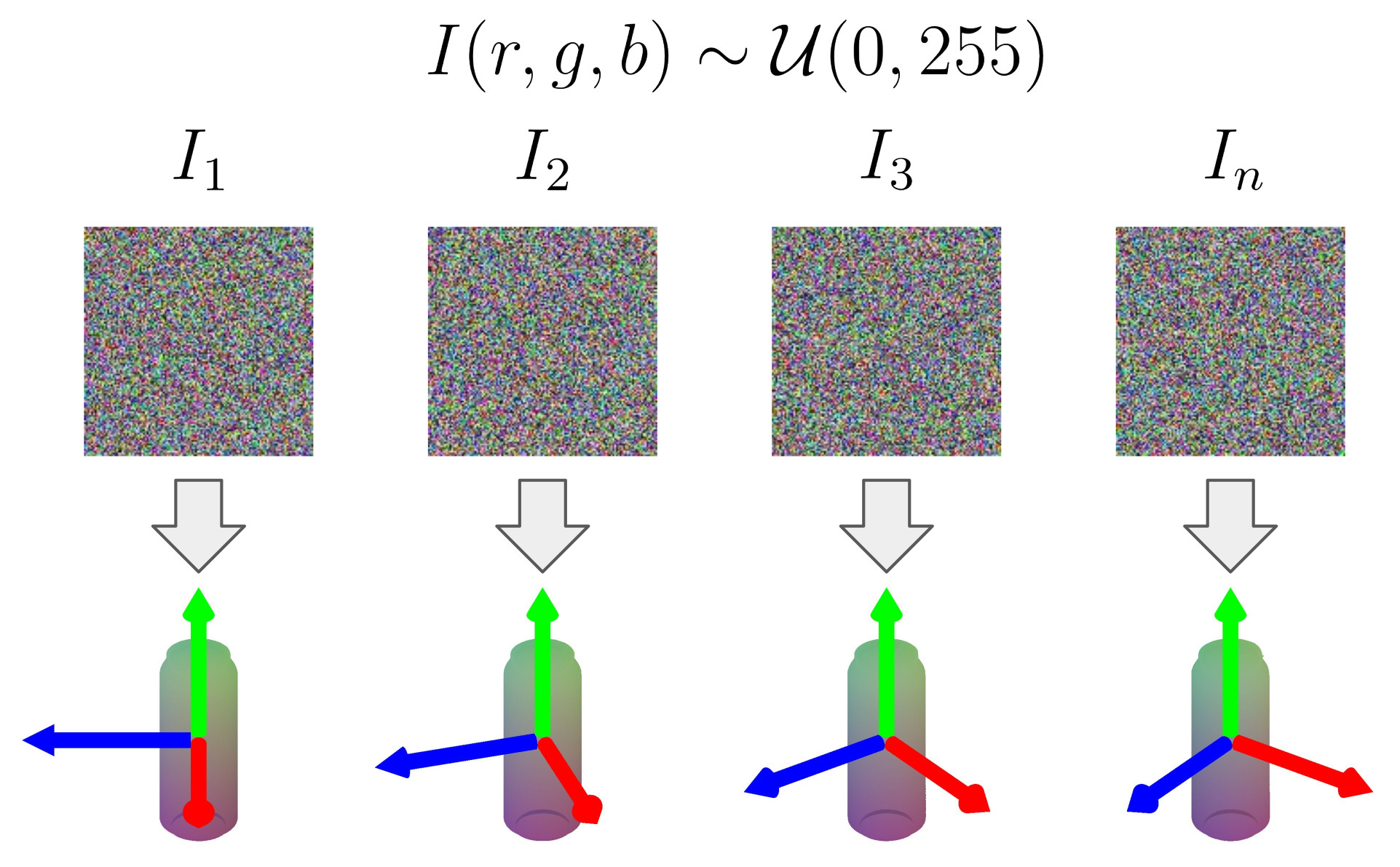}
   \caption{\textbf{Illustration of the probabilistic loss used by DiffusionNOCS.} For each sample around the symmetry axis r,g,b values from a uniform Gaussian distribution are sampled. This introduces an adversarial character to the NOCS prediction, prohibiting the network from converging to local minima that do not capture the full symmetry of the object. By introducing this adverarial effect the network learns the symmetry of the object implicitly.}
   \label{fig:diffusionnocs_loss}
\end{figure}

\cite{ikeda2024diffusionnocs} presents an implicit symmetry handling approach for learning pose probabilities by sampling an additional noise input to the diffusion model.
Compared to the other papers, they sample multiple noise inputs for each training sample drawn from the dataloader.
Consequently, each instance of the training set has newly generated noise samples in each epoch during training.
After predicting the NOCS the inlier rate of the TEASER++ pointcloud registration result is used as an additional loss for backpropagation.
An example of this noise sampling approach is illustrated in Fig.~\ref{fig:diffusionnocs_loss}.
The advantage of this type of symmetry handling is that no discrete symmetry annotations are necessary.

\section{EXPERIMENTS}
\label{sec:experiments}

\begin{table*}[t!]
\centering
\caption{\textbf{Evaluation on the CAMERA and REAL275 datasets.} mAP scores for 3D bounding boxes, $\mathbf{R}$ and $\mathbf{t}$. "-" indicates scores not reported in the respective papers. Best performance in \textbf{bold}, worst performance / RGB-only \underline{underlined}\\
\textsuperscript{\textdagger} use synthetically rendered training data, evaluate on real test data.}
\label{tab:pose_estimation_accuracy}
\begin{tabular}{c|c|c|c|c|ccc|ccc}
\toprule
\multirow{2}{*}{\textbf{Dataset}} & \multirow{2}{*}{\textbf{Year}} & \multirow{2}{*}{\textbf{Paper}} & \multirow{2}{*}{\textbf{Input}} & \multirow{2}{*}{\textbf{Priors}} & \multicolumn{6}{c}{\textbf{mAP}} 
 \\
 & & & & & ${3D_{25}}$ & ${3D_{50}}$ & ${3D_{75}}$ & $5^\circ 5cm$ & $10^\circ 5cm$ & $10^\circ 10cm$ \\
\midrule
\multirow{2}{*}{\textbf{CAMERA}} & 2019 & Wang et al.~\cite{wang2019normalized} & RGB-D & end-to-end & \textbf{91.1} & 83.9 & 69.5 & 40.9 & 64.6 & 65.1 \\
& 2020 & Tian et al.~\cite{tian2020shape}  & RGB-D & Mask R-CNN~\cite{matterport_maskrcnn_2017} & - & 93.2 & 83.1 & 59.0 & 81.5 & -\\
& 2021 & Lee et al.~\cite{lee2021noce} & RGB  & Mask R-CNN~\cite{matterport_maskrcnn_2017} & \underline{75.4} & \underline{32.4} & \underline{5.1} & - & - & \underline{19.2}\\
& 2021 & Chen et al.~\cite{chen2021fs} & D  & YOLOv3~\cite{redmon2018yolov3incrementalimprovement} & - & - & 85.2 & 62.0 & - & -\\
& 2022 & Wang et al.~\cite{wang2021recurrent} & RGB-D & Mask R-CNN~\cite{matterport_maskrcnn_2017} & - & \textbf{93.8} & 88.0 & 76.4 & 87.7 & -\\
& 2023 & Zhang et al.~\cite{zhang2022ssp} & D &  Mask R-CNN~\cite{matterport_maskrcnn_2017} & - & - & 86.8 & 75.5 & 87.4 & - \\
& 2023 & Wan et al.~\cite{wan2023socs} & D  & Mask R-CNN~\cite{matterport_maskrcnn_2017} & - & - & - & - & - & -\\
& 2023 & Wang et al.~\cite{wang2023query6dof} & D & Mask R-CNN~\cite{matterport_maskrcnn_2017} & - & 92.3 & 88.6 & \textbf{83.9} & \textbf{90.5} & -\\
& 2023& Zou et al.~\cite{zou2023gpt} & D &  Mask R-CNN~\cite{matterport_maskrcnn_2017} & - & 92.5 & 86.9 & 76.5 & 88.7 & \textbf{89.9} \\
& 2024 & Ikeda et al.~\cite{ikeda2024diffusionnocs} & RGB-D  & Mask R-CNN~\cite{matterport_maskrcnn_2017} & - & - & - & - & - & -  \\
& 2024 & Fan et al.~\cite{fan2024acrpose} & RGB-D & Mask R-CNN~\cite{matterport_maskrcnn_2017} & - & 93.7 & \textbf{89.6} & 75.1 & 89.5 & -\\
& 2024 & Krishnan et al.~\cite{krishnan2024omninocs} & RGB & Mask R-CNN~\cite{matterport_maskrcnn_2017} & - & - & - & - & - & -  \\

\midrule

\multirow{2}{*}{\textbf{REAL275}} & 2019 & Wang et al.~\cite{wang2019normalized} & RGB-D  & end-to-end & 84.9 & 80.5 & 30.1 & 10.0 & 26.7 & 26.7 \\
& 2020 & Tian et al.~\cite{tian2020shape} & RGB-D & Mask R-CNN~\cite{matterport_maskrcnn_2017} &  - & 77.3 & 53.2 & 21.4 & 54.1 & - \\
& 2021 & Lee et al.~\cite{lee2021noce} & RGB  & Mask R-CNN~\cite{matterport_maskrcnn_2017} & \underline{62.0} & \underline{23.4} & \underline{3.0} & - & - & \underline{9.6}\\
& 2021 & Chen et al.~\cite{chen2021fs} & D & YOLOv3~\cite{redmon2018yolov3incrementalimprovement} & \textbf{95.1} & \textbf{92.2} & 63.5 & 28.2 & 60.8 & 64.6\\
& 2021 & Wang et al.~\cite{wang2021recurrent} & RGB-D & Mask R-CNN~\cite{matterport_maskrcnn_2017} & - & 79.3 & 55.9 & 34.3 & 60.8 & -\\
& 2022 & Zhang et al.~\cite{zhang2022ssp} & D & Mask R-CNN~\cite{matterport_maskrcnn_2017} & 84.0 & 81.1 & 52.0 & 33.9 & 69.1 & 71.0\\
& 2023 & Wan et al.~\cite{wan2023socs} & D & Mask R-CNN~\cite{matterport_maskrcnn_2017} & - & 82.0 & 75.0 & \textbf{56.0} & \textbf{82.0} & -\\
& 2023 & Wang et al.~\cite{wang2023query6dof} & D & Mask R-CNN~\cite{matterport_maskrcnn_2017} & - & 82.9 & \textbf{76.0} & 54.7 & 81.6 & -\\
& 2023 & Zou et al.~\cite{zou2023gpt} & D & Mask R-CNN~\cite{matterport_maskrcnn_2017} & - & 82.0 & 70.4 & 53.8 & 77.7 & \textbf{79.8}\\
& 2024 & Ikeda et al.~\cite{ikeda2024diffusionnocs}\textsuperscript{\textdagger} & RGB-D & Mask R-CNN~\cite{matterport_maskrcnn_2017} & - & - & - & 35.0 & 66.6 & - \\
& 2024 & Fan et al.~\cite{fan2024acrpose} & RGB-D & Mask R-CNN~\cite{matterport_maskrcnn_2017} & - & 82.3 & 66.6 & 41.3 & 67.0 & -\\
& 2024 & Krishnan et al.~\cite{krishnan2024omninocs} & RGB & Mask R-CNN~\cite{matterport_maskrcnn_2017} & \underline{43.5} & \underline{10.6} & - & - & - & -  \\
\bottomrule
\end{tabular}
\end{table*}

\begin{figure}[t]
   \centering
    \includegraphics[width=0.8\columnwidth]{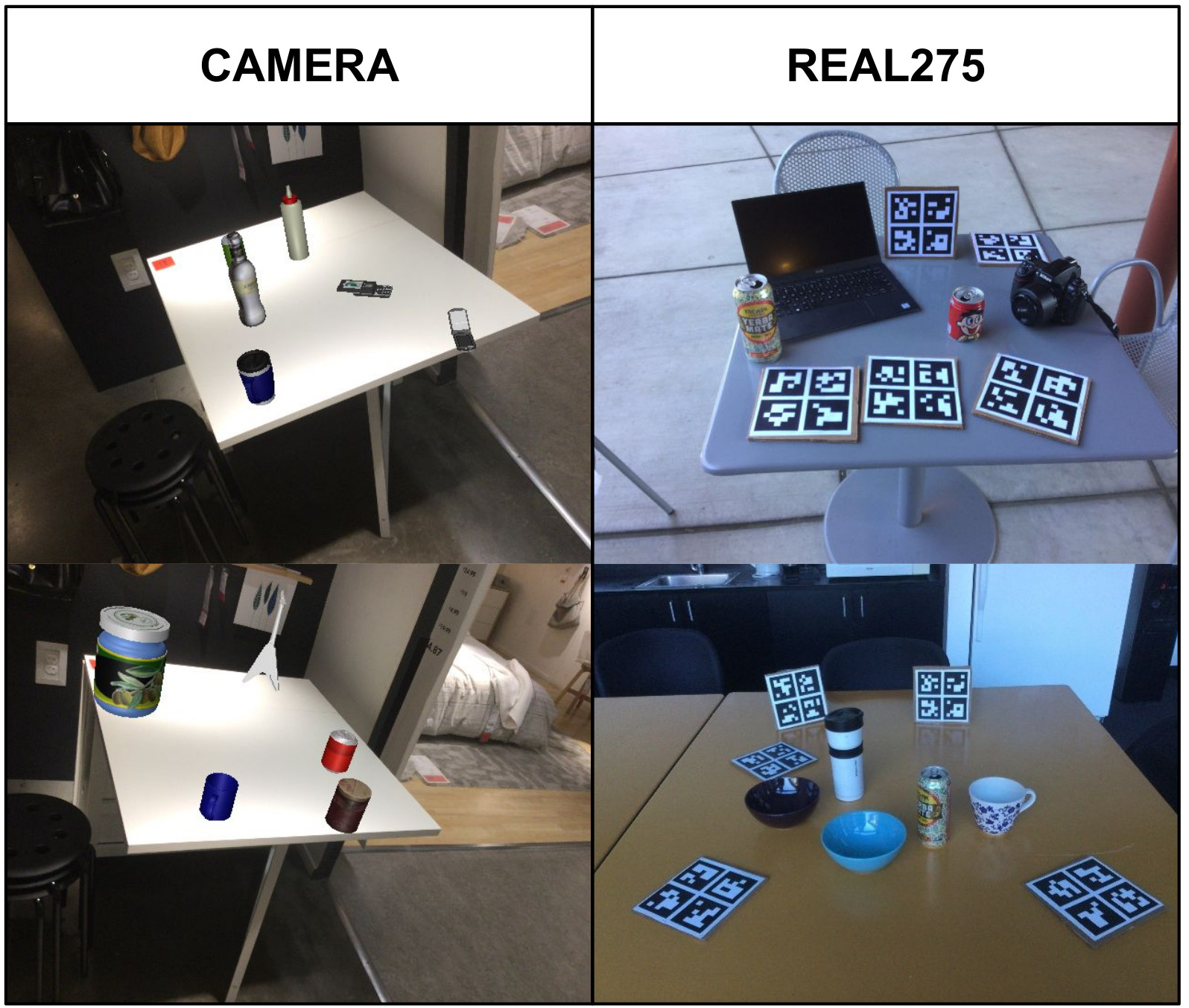}
   \caption{\textbf{Exemplary images of the CAMERA and REAL275 datasets.} While the CAMERA dataset features rendered object instances on real backgrounds, the REAL275 dataset solely features real image data.}
   \label{fig:datasets}
\end{figure}

The selected papers report results on the REAL275 and Context-Aware MixEd ReAlity (CAMERA) datasets.
Both datasets were introduced in~\cite{wang2019normalized}.
Exemplary images are shown in Fig.~\ref{fig:datasets}.
The CAMERA dataset consists of 300k synthetically rendered object images pasted onto real backgrounds.
The REAL275 dataset contains 2750 test images across 18 different scenes.
Both datasets contain the same object instances of 6 different object categories (bottle, bowl, camera, can, laptop, and mug), consisting of non-rotation-symmetric (camera, laptop, mug) and continuously symmetric (bottle, bowl, can) objects.
Since the CAMERA dataset also uses synthetic object renderings for the test data, its relevance for evaluating algorithms for real-world pose estimation capabilities is questionable.
If the test data is synthetic, the performance of algorithms on the real-world domain can not be evaluated.

Regarding accuracy metrics, the authors of \cite{wang2019normalized, tian2020shape, lee2021noce, krishnan2024omninocs} report the mean Average Precision (mAP) of $3D_{25}$, $3D_{50}$, and $3D_{75}$ scores, which represent the Intersection over Union (IoU) between ground truth and estimated 3D bounding box at 25\%, 50\%, and 75\% respectively.
The authors of \cite{wang2019normalized, tian2020shape, lee2021noce, ikeda2024diffusionnocs} papers also report the mAP below error thresholds of $\mathbf{R}$ and $\mathbf{t}$.
The specific thresholds in the papers vary.
Therefore, the most common ones are selected ($5^\circ 5cm$, $10^\circ 5cm$, and $10^\circ 10cm$).

\section{RESULTS}
\label{sec:results}
Table II shows the results presented in the selected papers on the CAMERA and REAL275 datasets.
Out of all, the two papers that use solely RGB data,~\cite{lee2021noce, krishnan2024omninocs} perform worst.
This confirms the hypothesis that regressing object size and translation without depth data during runtime is challenging.
While~\cite{lee2021noce} report 75.4\% of 3D bounding boxes within an IoU of 25\%, the numbers drop to 32.4\% at an IoU of 50\% and 5.1\% at an IoU of 75\%.
An opposite trend can be observed for the methods using solely depth.
The best performing methods for 3D bounding box and $\mathbf{R}$, $\mathbf{t}$ prediction rely on depth only.
Especially \cite{wan2023socs, wang2023query6dof} appear to be the best performing methods on the REAL275 dataset for predicting $\mathbf{R}$, $\mathbf{t}$.
\cite{wan2023socs} is slightly in the lead with 56.0\% mAP $5^\circ 5cm$ and 82.0\% $10^\circ 5cm$.
\cite{wang2023query6dof} is close with 54.7\% mAP $5^\circ 5cm$ and 81.6\% $10^\circ 5cm$.
Both \cite{wan2023socs, wang2023query6dof} do not perform explicit symmetry handling.
While performance metrics for individual object categories are missing in the papers, it appears that the 3D GCN of \cite{wan2023socs} and the CNNs + MLP network of \cite{wang2023query6dof} handle the symmetries of objects well enough.
Both papers also use learning-based 6D pose solving, namely anisotropic scaling~\cite{wan2023socs} and direct regression~\cite{wang2023query6dof}.
This sophisticated 6D pose solving techniques could be the reason for the superior performances.
On the other hand, ~\cite{zou2023gpt} comes close to the results of \cite{wan2023socs, wang2023query6dof} while employing the deterministic Umeyama method~\cite{umeyama1991least} for rigid point cloud alignment.
The probabilistic loss of \cite{ikeda2024diffusionnocs} does not lead to improved performance.
However, since the authors of~\cite{ikeda2024diffusionnocs} use synthetic renderings for training and real data for evaluation, a fair comparison is not possible.

When comparing pose estimation performance, the source of object detection priors have to be taken into account.
While \cite{tian2020shape, lee2021noce, wang2021recurrent, zhang2022ssp, wan2023socs, wang2023query6dof, zou2023gpt, ikeda2024diffusionnocs, fan2024acrpose, krishnan2024omninocs} use pre-computed Mask R-CNN location priors to ensure fair comparison,~\cite{wang2019normalized} use the location priors of their end-to-end trainable network and~\cite{chen2021fs} use YOLOv3 for location priors.
This results in superior results for 3D bounding box estimation on REAL275 by ~\cite{chen2021fs}.

Overall, results are better on the CAMERA dataset compared to the REAL275 dataset.
This can likely be attributed to the stronger domain shift between the training and test data of REAL275.
Not only objects but also scenes are different, including other lighting, shadow and reflection.
The domain shift for pose estimation only consists of object difference between training and test dataset.

Lastly, a clear correlation between improved performance and newer network architectures such as transformers or diffusion models cannot be observed.
% A development of category-level object pose estimation performance is illustrated in Fig.~\ref{fig:real275}.
% The graphs depict a spline interpolation based on the mAP scores between 2019 and 2024 on the REAL275 dataset.
% Performance improved between 2019 to 2023 with a slight decline in 2024.

% \begin{figure}[t]
%    \centering
%     \includegraphics[width=1.0\columnwidth]{real275_plot.png}
%    \caption{\textbf{Pose Estimation Performance on REAL275 from 2019 to 2024.}}
%    \label{fig:real275}
% \end{figure}

\section{CONCLUSION}
\label{sec:conclusion}
This paper compares category-level object pose estimation methods which are evaluated on the CAMERA and REAL275 datasets.
The methods differ regarding input modalities, symmetry handling, network types, and 6D pose solver algorithms.
A comprehensive comparison was conducted, focusing on symmetry handling and its potential impact on model performance. 
After reviewing input modalities, network architectures, 6D pose solvers, symmetry handling, experiments, and results, the following conclusions are drawn:

\begin{itemize}
    \item Omitting depth as done by~\cite{lee2021noce, krishnan2024omninocs} drastically reduces performance as compared to RGB and RGB-D based methods.
    \item While the absence of depth data worsens results, the depth-only methods perform best overall, indicating that depth data is crucial for improving category-level object pose estimation. 
    \item The two best performing methods use no explicit symmetry handling, suggesting that implicit symmetry handling is not mandatory if model architecture is allowing for it.
    \item While methods with learning-based 6D pose solvers excel regarding pose estimation performance, papers using deterministic methods such as Umeyama~\cite{umeyama1991least} achieve results almost on par.
    This is crucial since bridging category-level and open-set pose estimation benefits from deterministic 3D geometry-agnostic algorithms for 6D pose solving.
    \item The usage of different 2D object detection priors hinders a fair comparison, since improvement cannot clearly attributed to either detection or pose estimation.
\end{itemize}

Future research should build upon this comparative paper by analyzing the individual aspects of category-level object pose estimation further.

% This command serves to balance the column lengths on the last page of the 
% document manually. It shortens  the textheight of the last page by a   
% suitable amount. This command does not take effect until the  next page so it 
% should come on the page before the  last. Make sure that you do not shorten 
% the textheight too much. 
\addtolength{\textheight}{-12cm}   

\section*{ACKNOWLEDGMENT}
This research is supported by the EU program EC Horizon 2020 for Research and Innovation under grant agreement No. 101017089, project TraceBot, and the Austrian Science Fund (FWF), under project No. I 6114, iChores.

\bibliographystyle{IEEEtran} % We choose the "plain" reference style
\bibliography{bib} % Entries are in the refs.bib file

\end{document}